\documentclass{article}
\usepackage[final]{colm2026_conference}

\usepackage{microtype}
\usepackage{graphicx}
\usepackage{booktabs}
\usepackage{lineno}
\usepackage{hyperref}
\usepackage{url}
\usepackage{amsmath}
\usepackage{amssymb}
\usepackage{amsthm}
\usepackage[capitalize,noabbrev]{cleveref}
\usepackage[ruled,vlined]{algorithm2e}
\usepackage{pgfplots}
\pgfplotsset{compat=1.17}
\usepgfplotslibrary{fillbetween}
\usepackage{tikz}

\makeatletter
\renewcommand\paragraph{\@startsection{paragraph}{4}{\z@}%
  {1.2ex \@plus .3ex \@minus .2ex}%
  {-0.6em}%
  {\normalfont\normalsize\bfseries}}
\makeatother

\definecolor{darkblue}{rgb}{0, 0, 0.5}
\hypersetup{colorlinks=true, citecolor=darkblue,
            linkcolor=darkblue, urlcolor=darkblue}

\definecolor{myblue}  {RGB}{0,114,178}
\definecolor{myorange}{RGB}{230,159,0}
\definecolor{mygreen} {RGB}{0,158,115}
\definecolor{myred}   {RGB}{213,94,0}

\title{RLVR Sharpens the Pencil, Not the Mind%
\thanks{Accepted at the Workshop on Methods and Opportunities at Small Scale
(MOSS), co-located with COLM~2026.}}

\author{Shreeyam Bangera \quad Manya Gupta\\
Department of Physics, Indian Institute of Technology Roorkee\\
\texttt{shreekant\_sb@ph.iitr.ac.in} \quad \texttt{manya\_g@ph.iitr.ac.in}}

\begin{document}

\ifcolmsubmission
\linenumbers
\fi

\date{}
\maketitle

\begin{abstract}
Models like DeepSeek-R1 made Reinforcement Learning with Verifiable Rewards
(RLVR) famous: reward a model for correct answers and it gets dramatically
better at reasoning without being taught how. The leading explanation is
\textit{sharpening}---that the base model already knows how to solve these
problems and RLVR just makes it more likely to reach for its best solution
paths. This has only ever been checked by comparing before and after training,
never by watching what changes inside across it. We tracked OLMo-2-1B across
all 13 training checkpoints on GSM8K. First-try accuracy climbs from 58.0\%
to 61.3\%, but ask the model 256 times and solution coverage actually rises
from 94.5\% to 96.9\%, proving the knowledge was already there. We then
planted reasoning mistakes into problems and trained simple probes to detect
them. Those probes held between AUROC 0.798 and 0.840 across all 13
checkpoints and 3 error types, a spread smaller than the confidence interval;
the model never builds a sharper internal picture of what a reasoning mistake
looks like. The one place we find real internal change is at the final answer
token, where a probe distinguishing early from late checkpoints hits an AUROC
of 0.93, while the same probe on the actual reasoning steps lands near chance.
RLVR makes the model lead with its best thinking. It does not change what that
thinking is.
\end{abstract}

\section{Introduction}

Models like DeepSeek-R1 and OpenAI o1 are trained by rewarding correct answers
with no supervision over how they get there, and show large gains on math and
reasoning benchmarks \citep{deepseek2025,openai2024}. The field has largely
settled on an explanation: sharpening. The base model already knows how to
solve these problems; RLVR just makes it more likely to reach for its best
solution paths \citep{yue2025,karan2025}. This is supported both by
behavioral evidence, showing that solutions are reachable from the base model
given enough attempts \citep{yue2025}, and mechanistic evidence, showing that
reasoning directions pre-exist in base model activations before any RLVR
training \citep{venhoff2025}. But the behavioral work tracks outputs across
training checkpoints while the mechanistic work only compares a base model to
a trained model at a single endpoint. Nobody has asked whether internal
representations of reasoning actually change as training progresses.

We track OLMo-2-1B across all 13 RLVR checkpoints on grade-school arithmetic,
measuring both outputs and internal states at each step. Unlike the
boundary-narrowing Yue et al.\ find in larger models, pass@256 does not shrink
here, giving us a clean window into whether representational change accompanies
behavioral change. It does not. Internal representations of reasoning errors
hold flat across all 13 checkpoints, while the only detectable change
concentrates at the final answer token rather than at the reasoning steps where
errors occur. We report a single-model, single-task (arithmetic) result and
scope claims accordingly.

\section{Related Work}

\paragraph{Behavioral evidence for sharpening.}
Yue et al.\ track pass@$k$ up to $k{=}256$ across training checkpoints and
find that RLVR improves pass@1 while narrowing the set of problems the model
can eventually solve, concluding that the capability was already there in the
base model \citep{yue2025}. Karan and Du reach the same conclusion without any
training at all, recovering RLVR-level performance through likelihood-guided
resampling from the base model \citep{karan2025}. Dang et al.\ offer a
mode-collapse account of why the boundary narrows \citep{dang2025}. These
works map how outputs change across training; what they leave open is whether
internal representations shift in parallel.

\paragraph{Mechanistic evidence for sharpening.}
Venhoff et al.\ use sparse autoencoders and steering vectors to show that
reasoning directions already exist in base model activations, with
post-training teaching the model when to use them rather than creating them
\citep{venhoff2025}. This is an endpoint comparison between a base model and a
trained model. We ask the same question but across all training checkpoints,
and separate it into two parts: do outputs change, and do internal
representations of reasoning change.

\paragraph{What changes during RLVR training.}
Recent work finds that RLVR drives near-linear drift in weights and outputs
\citep{wang2026} and that distributional shifts are sparse, close to zero at
most token positions and large at a few \citep{meng2026}. Yan et al.\ show
that representational change, where it exists, can be localized causally
\citep{yan2026}. We test whether arithmetic error representations fall in the
invariant majority or the shifting minority.

\paragraph{Error representations in activations.}
Linear probes can detect truthfulness and error states in model activations
\citep{azaria2023}, but such signals can be diagnostic rather than causal
\citep{yuan2026} and can reflect surface features rather than genuine reasoning
\citep{haller2025}. We control for this directly in
Section~\ref{sec:invariant}.

\section{Setup}

\textbf{Model.} OLMo-2-0425-1B-Instruct \citep{olmo2025}; 16 layers
($d{=}2048$); 13 checkpoints at steps 200--2600 (interval 200).

\textbf{Task.} GSM8K \citep{cobbe2021} test split; 1319 problems;
step-by-step generation; final numeric answer parsed.

We measure how well the model solves problems at each checkpoint and whether
its internal states encode reasoning errors differently as training progresses,
via two procedures (full pseudocode in Appendix~\ref{sec:algorithms}): a
\emph{behavioral} sweep that samples completions per checkpoint to compute
pass@$k$ at both $K{=}32$ (wide, all checkpoints) and $K{=}256$ (deep,
5-checkpoint subset); and a \emph{representational} probe that injects one of
three controlled arithmetic errors into correctly solved problems and trains
logistic-regression probes on PCA-reduced hidden states at six readout sites
to test whether checkpoints differ in how those errors are encoded.

\textbf{Code.} All code, per-checkpoint probe outputs, and figure-generation
scripts are released at
\url{https://github.com/Leviathan2006/rlvr-pencil}.

\section{Results}

\begin{figure}[t]
\centering
\begin{minipage}{0.48\linewidth}
\centering
\begin{tikzpicture}
\begin{axis}[
  width=\linewidth, height=4.0cm,
  xlabel={RLVR step}, ylabel={pass@$k$},
  ymin=0.55, ymax=1.02, xmin=100, xmax=2700,
  legend style={
    font=\scriptsize, at={(0.5,-0.36)}, anchor=north,
    legend columns=4, draw=none, column sep=0.4em},
  tick label style={font=\scriptsize},
  label style={font=\small},
  title style={font=\small},
  title={(a) Coverage across training ($K{=}256$)},
  grid=major, grid style={gray!20},
]
\addplot[mark=*,        myblue,   thick] coordinates
  {(200,0.597)(800,0.606)(1400,0.625)(2000,0.628)(2600,0.626)};
\addplot[mark=triangle*,myorange, thick] coordinates
  {(200,0.834)(800,0.838)(1400,0.846)(2000,0.842)(2600,0.842)};
\addplot[mark=square*,  mygreen,  thick] coordinates
  {(200,0.895)(800,0.890)(1400,0.900)(2000,0.895)(2600,0.898)};
\addplot[mark=diamond*, myred,    thick] coordinates
  {(200,0.945)(800,0.958)(1400,0.961)(2000,0.962)(2600,0.969)};
\legend{pass@1, pass@8, pass@32, pass@256}
\end{axis}
\end{tikzpicture}
\end{minipage}\hfill
\begin{minipage}{0.48\linewidth}
\centering
\begin{tikzpicture}
\begin{axis}[
  width=\linewidth, height=4.0cm,
  xlabel={RLVR step},
  ylabel={$\Delta$ from step 200 ($\times 10^{-2}$)},
  xmin=100, xmax=2700, ymin=-0.2, ymax=3.2,
  legend style={
    font=\scriptsize, at={(0.5,-0.36)}, anchor=north,
    legend columns=2, draw=none, column sep=0.4em},
  tick label style={font=\scriptsize},
  label style={font=\small},
  title style={font=\small},
  title={(b) pass@1 rises, deep coverage holds},
  grid=major, grid style={gray!20},
]
\addplot[mark=*,        myblue, thick] coordinates
  {(200,0.0)(800,0.9)(1400,2.8)(2000,3.1)(2600,2.9)};
\addplot[mark=diamond*, myred,  thick] coordinates
  {(200,0.0)(800,1.3)(1400,1.6)(2000,1.7)(2600,2.4)};
\draw[gray, dashed] (axis cs:100,0) -- (axis cs:2700,0);
\legend{$\Delta$ pass@1, $\Delta$ pass@256}
\end{axis}
\end{tikzpicture}
\end{minipage}
\caption{First-try accuracy and deep coverage across training.
(a)~pass@$k$ curves from the deep sweep.
(b)~pass@1 rises by $+3.3$ pts while pass@256 does not narrow;
the model's solution coverage stays intact across training.}
\label{fig:passk}
\end{figure}
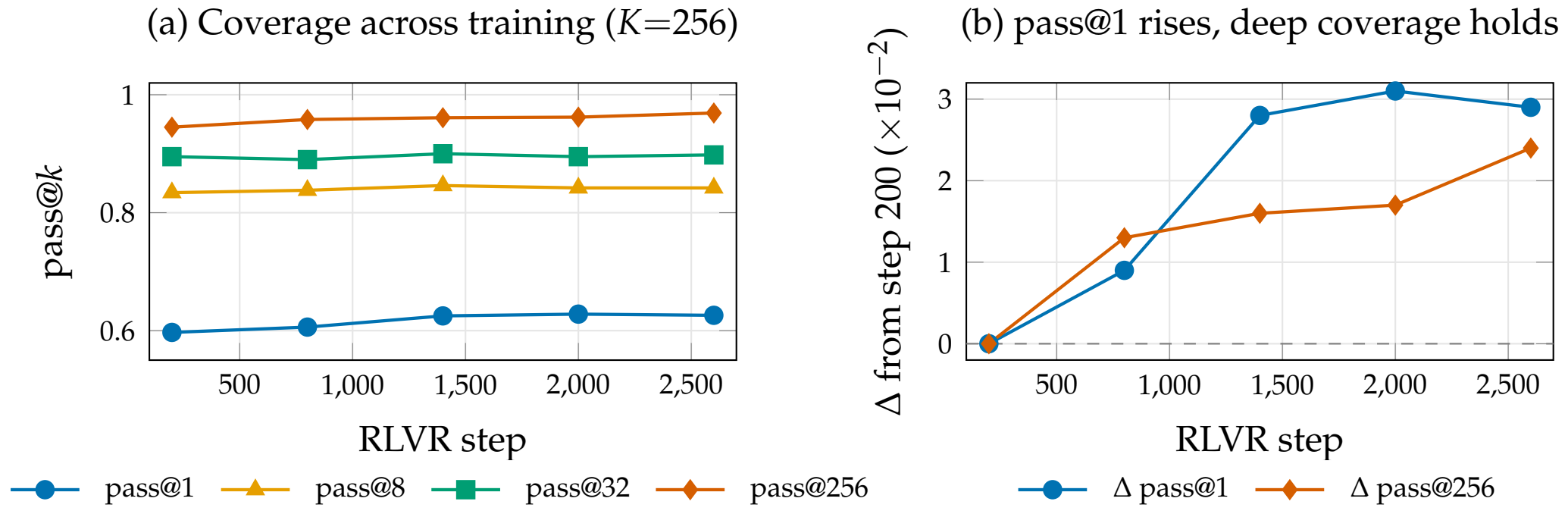

\subsection{Behavior Becomes More Precise Without Losing Breadth}

pass@1 rises from 0.580 to 0.613 across training. The apparent drop in
pass@32 is an artifact of small $k$: pass@256 actually rises from 0.945
to 0.969 (\cref{fig:passk}), so the set of problems the model can eventually
solve grows, not shrinks. This differs from the boundary-narrowing Yue et
al.\ and Dang et al.\ find in larger models \citep{yue2025,dang2025}, and
gives us a clean window into whether internal representations shift
alongside behavior.

\subsection{Internal Error Representations Remain Unchanged}
\label{sec:invariant}

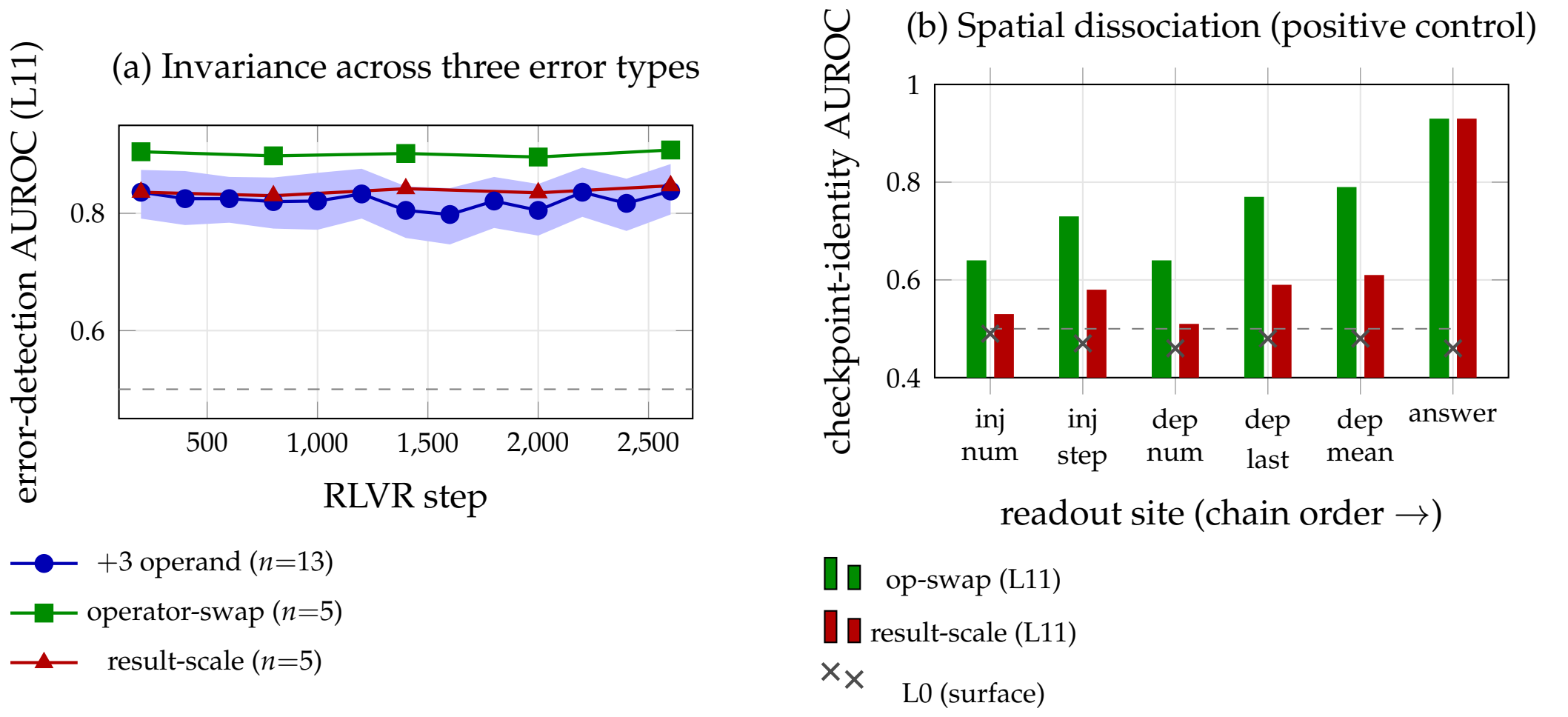
\begin{figure}[t]
\centering
\begin{minipage}{0.48\linewidth}
\centering
\begin{tikzpicture}
\begin{axis}[
  width=\linewidth, height=4.2cm,
  xlabel={RLVR step},
  ylabel={error-detection AUROC (L11)},
  xmin=100, xmax=2700, ymin=0.45, ymax=0.95,
  legend to name=legA,
  legend style={font=\scriptsize, draw=none, fill=none,
                legend columns=1, row sep=2pt},
  tick label style={font=\scriptsize},
  label style={font=\small},
  title style={font=\small},
  title={(a) Invariance across three error types},
  grid=major, grid style={gray!20},
]
\addplot[name path=hi, draw=none, forget plot] coordinates {
  (200,0.874)(400,0.872)(600,0.862)(800,0.861)(1000,0.869)
  (1200,0.876)(1400,0.846)(1600,0.843)(1800,0.862)(2000,0.850)
  (2200,0.878)(2400,0.859)(2600,0.884)};
\addplot[name path=lo, draw=none, forget plot] coordinates {
  (200,0.791)(400,0.780)(600,0.784)(800,0.774)(1000,0.772)
  (1200,0.791)(1400,0.758)(1600,0.747)(1800,0.775)(2000,0.762)
  (2200,0.794)(2400,0.770)(2600,0.798)};
\addplot[blue!25, forget plot] fill between[of=hi and lo];
\addplot[mark=*, color=blue!70!black, thick] coordinates {
  (200,0.836)(400,0.825)(600,0.825)(800,0.820)(1000,0.821)
  (1200,0.833)(1400,0.805)(1600,0.798)(1800,0.821)(2000,0.805)
  (2200,0.836)(2400,0.817)(2600,0.838)};
\addlegendentry{$+3$ operand ($n{=}13$)}
\addplot[mark=square*, color=green!55!black, thick] coordinates {
  (200,0.905)(800,0.898)(1400,0.902)(2000,0.896)(2600,0.908)};
\addlegendentry{operator-swap ($n{=}5$)}
\addplot[mark=triangle*, color=red!70!black, thick] coordinates {
  (200,0.836)(800,0.830)(1400,0.842)(2000,0.835)(2600,0.847)};
\addlegendentry{result-scale ($n{=}5$)}
\draw[gray, dashed] (axis cs:100,0.5) -- (axis cs:2700,0.5);
\end{axis}
\end{tikzpicture}
\smallskip
\raggedright\pgfplotslegendfromname{legA}
\end{minipage}\hfill
\begin{minipage}{0.48\linewidth}
\centering
\begin{tikzpicture}
\begin{axis}[
  width=\linewidth, height=4.2cm,
  ybar, bar width=5pt,
  enlarge x limits=0.12,
  xlabel={readout site (chain order $\rightarrow$)},
  ylabel={checkpoint-identity AUROC},
  symbolic x coords={injnum,injstep,depnum,deplast,depmean,answer},
  xtick=data,
  xticklabels={
    \shortstack{inj\\num}, \shortstack{inj\\step},
    \shortstack{dep\\num}, \shortstack{dep\\last},
    \shortstack{dep\\mean}, answer},
  xticklabel style={align=center, font=\scriptsize},
  ymin=0.4, ymax=1.0,
  legend to name=legB,
  legend style={font=\scriptsize, draw=none, fill=none,
                legend columns=1, row sep=2pt},
  tick label style={font=\scriptsize},
  label style={font=\small},
  title style={font=\small},
  title={(b) Spatial dissociation (positive control)},
  grid=major, grid style={gray!20},
]
\addplot[fill=green!55!black, draw=none] coordinates {
  (injnum,0.64)(injstep,0.73)(depnum,0.64)
  (deplast,0.77)(depmean,0.79)(answer,0.93)};
\addlegendentry{op-swap (L11)}
\addplot[fill=red!70!black, draw=none] coordinates {
  (injnum,0.53)(injstep,0.58)(depnum,0.51)
  (deplast,0.59)(depmean,0.61)(answer,0.93)};
\addlegendentry{result-scale (L11)}
\addplot[only marks, mark=x, mark size=3pt,
         color=gray!60!black, line width=0.8pt] coordinates {
  (injnum,0.49)(injstep,0.47)(depnum,0.46)
  (deplast,0.48)(depmean,0.48)(answer,0.46)};
\addlegendentry{L0 (surface)}
\draw[gray, dashed] (axis cs:injnum,0.5) -- (axis cs:answer,0.5);
\end{axis}
\end{tikzpicture}
\smallskip
\raggedright\pgfplotslegendfromname{legB}
\end{minipage}
\caption{%
(a)~Error-detection AUROC at the dependent-step token (layer~11) across the
RLVR trajectory; shaded band is the bootstrap 95\,\% CI for the
$+3$-operand condition.
(b)~Checkpoint-identity AUROC by readout site; the L0 baseline ($\times$)
is near chance throughout.}
\label{fig:rep}
\end{figure}

For the $(+3)$ operand error, AUROC spans $0.798$--$0.840$ across all 13
checkpoints---a spread of $0.042$, smaller than the CI half-width of
${\sim}0.044$. The other two types hold the same pattern: operator-swap
(spread $0.032$, $R^2{=}0.005$) and result-scale (spread $0.033$,
$R^2{=}0.345$); \cref{fig:rep}(a). Drift is statistically
indistinguishable from zero across all three.

\paragraph{The signal is computed, not surface.}
Probes can pick up shallow cues like token length or activation norm rather
than anything about the reasoning itself \citep{haller2025}. A norm proxy
rules this out here, sitting at chance throughout ($0.50$--$0.52$). The probe
signal rises from chance at layer 0 (AUROC $0.498$) to a mid-layer peak of
$0.833$, with near-identical layer profiles across all 13 checkpoints. The
error information the model encodes is a mid-layer computation, not a surface
cue.

\paragraph{Positive control.}
Using the identical pipeline, we probe whether clean activations can
distinguish early from late checkpoints. At the output token, the probe
reaches AUROC $0.93$ for both operator-swap and result-scale, with the
layer-0 baseline at chance. At the intermediate reasoning tokens, it lands
near chance (\cref{fig:rep}(b)). The probe finds real training-induced change
at the output and finds none at the reasoning steps. The flat trajectory on
error representations is meaningful precisely because the method demonstrably
detects change where change exists.

\subsection{The Invariance Is Robust, Not an Artifact of Analysis Choices}
\label{sec:robustness}

The result holds across both pre-registered choices. Probing all 17 layers,
the cross-checkpoint spread stays within noise at 16 of 17; the per-checkpoint
layer profiles are near-identical in shape across training. Probing all six
readout sites, every site is invariant (spread $\leq 0.034$). Full tables and
figures are in \cref{sec:layersweep,sec:siterobust}.

\subsection{Training Alters the Final Output, Not the Intermediate Reasoning}

Bringing the two findings together: error-detection decodability is flat
across RLVR training ($\approx$constant, three error types, and robust to layer
and site choice; \cref{sec:robustness}), while first-try accuracy improves and
RLVR's representational footprint is large at the output token. In this setting,
RLVR acts on the output distribution rather than on the internal arithmetic
reasoning-error representation. A practical consequence is that probes or
verifiers trained to detect reasoning errors in the base model should transfer
across RLVR checkpoints, amortizing verification compute.

\section{Computational Budget and Reproducibility}
\label{sec:compute}
All experiments are inference-only: we evaluate publicly released OLMo-2-1B
checkpoints \citep{olmo2025} without updating any weights, using the standard
$2N$-FLOPs-per-token approximation ($N{=}10^9$, prompt ${\approx}150$ tokens,
generated chain ${\approx}180$ tokens). Total cost across all reported
experiments---behavioral pass@$k$ sweeps, representational probing across
three error types, and the layer- and site-robustness ablations---is
${\approx}\,2.0\times10^{17}$ FLOPs, roughly $465\times$ below the
$10^{20}$-FLOP single-run training cap; the behavioral sweeps dominate this
total since they generate up to 256 samples per problem. We release the
injection-and-probing pipeline, per-checkpoint activation tensors and probe
outputs (AUROC values with bootstrap CIs) for every layer, site, and error
type, and the pass@$k$ generation logs; the only external dependencies are
the public OLMo-2-1B checkpoints and the GSM8K test split \citep{cobbe2021}.

\section{Conclusion}
We establish that RLVR drives behavioral improvements by altering the final output distribution rather than reshaping internal arithmetic reasoning-error representations. A detailed discussion of our findings, along with the study's limitations and broader impacts, is provided in Appendix~\ref{sec:limitations}.

\appendix

\section{Detailed Procedures}
\label{sec:algorithms}

The two procedures summarized in Section~\ref{sec:invariant} are given here
in full.

\begin{algorithm}[H]
\caption{Behavioral Evaluation}
\KwIn{Checkpoints $\mathcal{C} = \{c_{200},\ldots,c_{2600}\}$,
      problems $\mathcal{P}$, temperature $\tau{=}0.8$}
\For{each checkpoint $c \in \mathcal{C}$}{
    Sample $K$ completions per problem at temperature $\tau$\;
    Compute pass@$k$ via unbiased estimator \citep{cobbe2021}\;
}
\tcp{\textbf{Wide sweep:} $K{=}32$, $n{=}200$, all 13 checkpoints}
\tcp{\textbf{Deep sweep:} $K{=}256$, $n{=}128$, 5-checkpoint subset}
\end{algorithm}

\begin{algorithm}[H]
\caption{Representational Probe}
\KwIn{Correctly solved problems $\mathcal{P}^{+}$, error types $\mathcal{E}$:
      operand $(+3)$, operator-swap, result-scale}
\For{each $p \in \mathcal{P}^{+}$, each $e \in \mathcal{E}$}{
    Inject $e$ into an early reasoning step; regenerate remainder\;
    Label chain \textsc{repair} if answer correct, \textsc{carry} if
    answer flips\;
    Extract hidden states at 6 readout sites: corrupted step,
    dependent step, answer token\;
}
\For{each readout site, each layer}{
    Fit PCA-256 on concatenated clean $+$ corrupt activations\;
    Train logistic regression; evaluate via 5-fold CV AUROC
    with bootstrap confidence intervals\;
}
\textbf{Pre-registered site:} layer 11, token \texttt{L\_dep\_num}\;
\end{algorithm}

\section{Discussion and Limitations}
\label{sec:limitations}

These results carry clear implications, and understanding their scope is important.

\textbf{(i) Complementary levels of analysis.}
pass@$k$ captures whole-problem solution-finding. The representational metric
is narrower: at one specific step, how do activations encode a single injected
arithmetic error? Behavioral evidence shows the output distribution tightening;
representational evidence shows the substrate encoding arithmetic errors is not
what changes. The two scales divide the question rather than compete over
it---and their combination localises where RLVR acts.

\textbf{(ii) Depth through focus.}
Arithmetic on OLMo-2-1B earns its specificity. Arithmetic gives unambiguous
ground truth for what counts as a reasoning error. Thirteen checkpoints provide
trajectory-level resolution across training. The constrained domain keeps
representational changes isolable---task variation would bury the within-training
dynamics we are trying to characterise. Scaling to diverse tasks and architectures
trades that granularity for coverage, and the signal goes with it.

\textbf{(iii) A clean window into RLVR.}
pass@1 rises 3.3 points; coverage does not narrow. That modesty is useful.
Without severe mode collapse, the main confound drops away---so the question
becomes clean: does representational change track behavioural change even when
the behavioural shift is real but small? Here, it does not. A noisier setting
would have buried that.

\textbf{(iv) A localized, not merely correlational, null.}
Linear probes establish that information is \emph{present} in activations but
not that the model uses it causally. We take this concern seriously. Our positive
control, however, directly addresses it: the identical probing pipeline
successfully and precisely detects where RLVR \emph{does} leave a
representational footprint---at the output token, with AUROC 0.93, rising
monotonically along the reasoning chain. The method demonstrably finds change
where change exists. Where it returns a flat trajectory, that null is meaningful.

The trajectory-level design fills a concrete gap left by representational
endpoint comparisons \citep{venhoff2025}, and situates one interpretable
arithmetic-error representation within the invariant majority that recent
sparse-shift work predicts \citep{meng2026}.

\section{Layer-Sweep Invariance}
\label{sec:layersweep}

\paragraph{Motivation.}
The main analysis pre-registers layer 11 as the reporting layer (median of best
layers across dry runs). A reviewer might ask whether the invariance result is
specific to that choice---whether other layers show drift that layer 11 happens
to miss. We address this by probing \emph{all 17 layers} (embedding through
layer 16) at five spanning checkpoints (steps 200, 800, 1400, 2000, 2600) using
the same injection pipeline and identical probe configuration ($n_{\text{repair}}
\approx 208$, PCA-256, 5-fold CV AUROC, bootstrap 95\% CIs).

\paragraph{Note on absolute AUROC.}
These ablation runs used a chain-collection target ($n_{\text{repair}}
+ n_{\text{carry}} \geq 300$) rather than the main analysis's fixed-attempt limit
(450 problems), yielding $n_{\text{repair}} \approx 208$ vs.\ $\approx 174$ in
the main analysis. The larger probe training set raises absolute AUROC by
approximately 0.10 relative to the main figures. The \emph{spread across
checkpoints}---our invariance measure---is unaffected by this, and is in fact
tighter here (mean spread 0.022 vs.\ 0.042 in the main analysis), providing
a stronger invariance signal.

\begin{figure}[h]
\centering
\begin{tikzpicture}
\begin{axis}[
  width=0.92\linewidth, height=5.0cm,
  xlabel={Layer}, ylabel={error-detection AUROC},
  xmin=-0.5, xmax=16.5, ymin=0.44, ymax=1.02,
  xtick={0,2,4,6,8,10,12,14,16},
  legend columns=5,
  legend style={font=\scriptsize, at={(0.5,1.14)}, anchor=south, draw=none,
    column sep=4pt},
  tick label style={font=\scriptsize}, label style={font=\small},
  title style={font=\small, yshift=1.0em},
  title={Layer-sweep AUROC profiles across five RLVR checkpoints},
  grid=major, grid style={gray!15},
]
\addplot[mark=*, mark size=1.3pt, color=blue!80!black, thick] coordinates {
  (0,0.500)(1,0.873)(2,0.952)(3,0.951)(4,0.973)(5,0.973)
  (6,0.987)(7,0.976)(8,0.980)(9,0.967)(10,0.960)
  (11,0.926)(12,0.920)(13,0.860)(14,0.851)(15,0.805)(16,0.805)};
\addplot[mark=triangle*, mark size=1.5pt, color=teal, thick] coordinates {
  (0,0.540)(1,0.857)(2,0.947)(3,0.945)(4,0.959)(5,0.966)
  (6,0.978)(7,0.967)(8,0.971)(9,0.959)(10,0.958)
  (11,0.930)(12,0.914)(13,0.872)(14,0.865)(15,0.833)(16,0.797)};
\addplot[mark=square*, mark size=1.3pt, color=orange!90!black, thick] coordinates {
  (0,0.502)(1,0.862)(2,0.951)(3,0.942)(4,0.962)(5,0.971)
  (6,0.983)(7,0.969)(8,0.977)(9,0.951)(10,0.961)
  (11,0.915)(12,0.903)(13,0.819)(14,0.807)(15,0.800)(16,0.785)};
\addplot[mark=diamond*, mark size=1.5pt, color=red!80!black, thick] coordinates {
  (0,0.523)(1,0.877)(2,0.954)(3,0.955)(4,0.972)(5,0.975)
  (6,0.984)(7,0.971)(8,0.976)(9,0.956)(10,0.964)
  (11,0.948)(12,0.936)(13,0.892)(14,0.865)(15,0.842)(16,0.839)};
\addplot[mark=pentagon*, mark size=1.3pt, color=purple!80!black, thick] coordinates {
  (0,0.522)(1,0.870)(2,0.953)(3,0.952)(4,0.965)(5,0.969)
  (6,0.981)(7,0.971)(8,0.972)(9,0.952)(10,0.954)
  (11,0.948)(12,0.936)(13,0.876)(14,0.856)(15,0.848)(16,0.827)};
\draw[gray!60, dashed, line width=0.6pt] (axis cs:-0.5,0.5) -- (axis cs:16.5,0.5)
  node[right, font=\scriptsize, gray] {chance};
\draw[gray!40, dotted, line width=0.6pt] (axis cs:11,-0.5) -- (axis cs:11,1.05)
  node[above, font=\scriptsize, gray] {L11};
\legend{step 200, step 800, step 1400, step 2000, step 2600}
\end{axis}
\end{tikzpicture}
\caption{Error-detection AUROC at every layer of OLMo-2-1B, for five RLVR
checkpoints. The five profiles are near-identical in shape: chance at layer 0,
rising to a peak around layer 6, declining in late layers. The vertical dotted
line marks the pre-registered reporting layer (L11). Near-identical profiles
across all five checkpoints confirm that the \emph{computational structure} of
the error representation---not just its endpoint decodability---is invariant
across RLVR training. Note: absolute AUROC is higher than in the main analysis
owing to a larger chain sample ($n_{\text{repair}} \approx 208$ vs.\ $\approx
174$); see text.}
\label{fig:layersweep}
\end{figure}
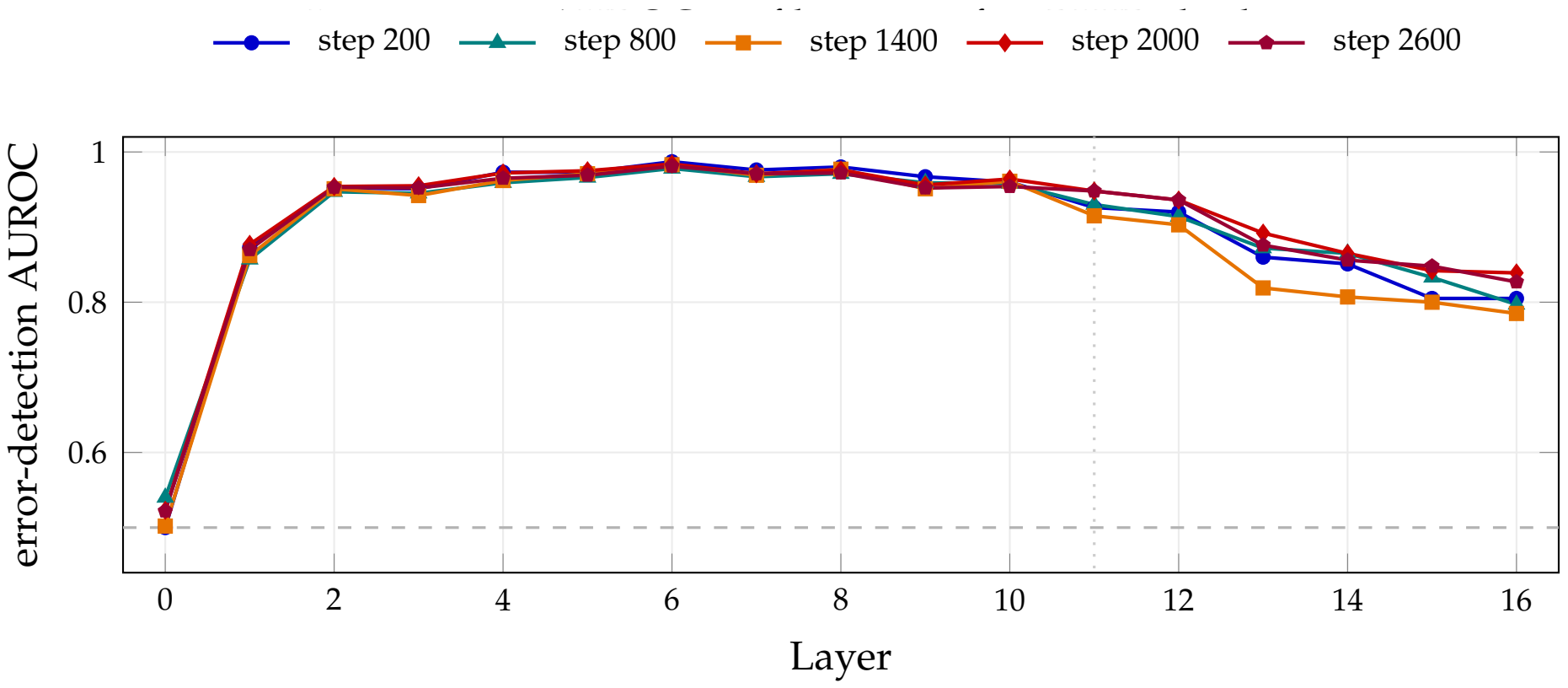

\paragraph{Results.}
Figure~\ref{fig:layersweep} shows the AUROC profile across all 17 layers for
each of the five checkpoints. Three observations stand out.

First, the layer-0 AUROC is at chance ($0.500$--$0.540$ across all checkpoints,
mean $0.517$), confirming the signal is a computed mid-layer representation, not
a surface artefact from token identity---consistent with the norm-proxy control
in the main paper.

Second, the five per-checkpoint profiles are \emph{near-identical in shape}:
all rise steeply from layer 0, peak around layer 6 (mean AUROC $0.983$), and
decline through the final layers. This shape-stability across training is
stronger evidence for representational invariance than a single-layer AUROC
comparison, since it shows the entire computational trajectory of the
representation is preserved.

Third, the cross-checkpoint spread is within noise at 16 of 17 layers. At the
peak (layer 6) the spread is only $0.008$, a fraction of the CI half-width
($0.009$). The one marginal exception is layer 13 (spread $0.073$, CI half-width
$0.031$), a late layer where absolute AUROC is already declining (mean $0.864$)
and signal is weaker. No systematic drift direction is visible there across
checkpoints. Table~\ref{tab:layersweep} reports the spread and CI half-width
for every layer.

\begin{table}[h]
\centering
\caption{Cross-checkpoint spread and mean CI half-width at each layer.
Spread $<$ 2$\times$CI half-width indicates invariance within noise; 16 of 17
layers satisfy this. Layer 13 is the marginal exception (spread 0.073, half-width
0.031) in a late layer where absolute signal is already declining.}
\label{tab:layersweep}
\small
\begin{tabular}{ccccc}
\toprule
Layer & Mean AUROC & Spread & CI half-width & Invariant \\
\midrule
0  & 0.517 & 0.039 & 0.050 & \checkmark \\
1  & 0.868 & 0.020 & 0.031 & \checkmark \\
2  & 0.952 & 0.007 & 0.019 & \checkmark \\
3  & 0.949 & 0.014 & 0.018 & \checkmark \\
4  & 0.966 & 0.014 & 0.013 & \checkmark \\
5  & 0.971 & 0.009 & 0.013 & \checkmark \\
6  & 0.983 & 0.008 & 0.009 & \checkmark \\
7  & 0.971 & 0.008 & 0.013 & \checkmark \\
8  & 0.975 & 0.009 & 0.011 & \checkmark \\
9  & 0.957 & 0.016 & 0.015 & \checkmark \\
10 & 0.959 & 0.010 & 0.015 & \checkmark \\
11 & 0.933 & 0.034 & 0.020 & \checkmark \\
12 & 0.922 & 0.033 & 0.023 & \checkmark \\
13 & 0.864 & 0.073 & 0.031 & $\times$ \\
14 & 0.849 & 0.058 & 0.035 & \checkmark \\
15 & 0.826 & 0.048 & 0.036 & \checkmark \\
16 & 0.811 & 0.054 & 0.036 & \checkmark \\
\bottomrule
\end{tabular}
\end{table}

\section{Readout-Site Invariance}
\label{sec:siterobust}

\paragraph{Motivation.}
The main analysis pre-registers one readout site (\texttt{L\_dep\_num}: the
number token in the step that depends on the corrupted step). A reviewer might
ask whether the invariance result is specific to that site, and whether other
readout positions---especially those closer to the injection point---show drift
that the pre-registered site misses. We probe all six readout sites at layer 11
across the same five checkpoints.

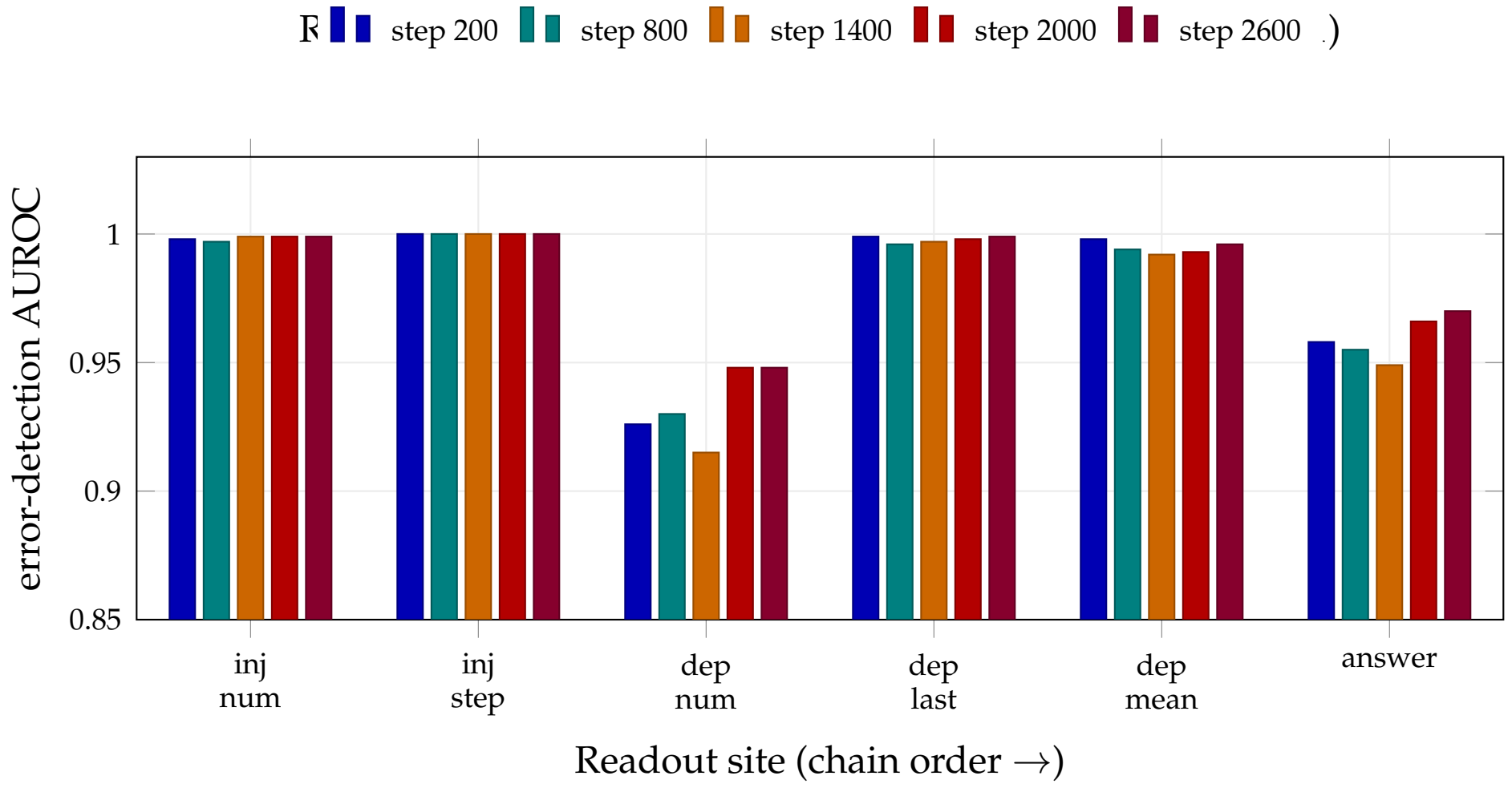
\begin{figure}[h]
\centering
\begin{tikzpicture}
\begin{axis}[
  width=0.92\linewidth, height=5.4cm,
  ybar, bar width=6pt,
  xlabel={Readout site (chain order $\rightarrow$)},
  ylabel={error-detection AUROC},
  symbolic x coords={injnum, injstep, depnum, deplast, depmean, answer},
  xtick=data,
  xticklabels={inj\\num, inj\\step, dep\\num, dep\\last, dep\\mean, answer},
  xticklabel style={align=center, font=\scriptsize},
  ymin=0.85, ymax=1.03,
  legend columns=5,
  legend style={font=\scriptsize, at={(0.5,1.20)}, anchor=south, draw=none,
    column sep=3pt},
  tick label style={font=\scriptsize}, label style={font=\small},
  title style={font=\small, yshift=1.5em},
  title={Readout-site AUROC across five RLVR checkpoints (layer 11)},
  grid=major, grid style={gray!15},
]
\addplot[fill=blue!70!black, draw=blue!50!black] coordinates {
  (injnum,0.998)(injstep,1.000)(depnum,0.926)(deplast,0.999)(depmean,0.998)(answer,0.958)};
\addplot[fill=teal, draw=teal!70!black] coordinates {
  (injnum,0.997)(injstep,1.000)(depnum,0.930)(deplast,0.996)(depmean,0.994)(answer,0.955)};
\addplot[fill=orange!80!black, draw=orange!60!black] coordinates {
  (injnum,0.999)(injstep,1.000)(depnum,0.915)(deplast,0.997)(depmean,0.992)(answer,0.949)};
\addplot[fill=red!70!black, draw=red!50!black] coordinates {
  (injnum,0.999)(injstep,1.000)(depnum,0.948)(deplast,0.998)(depmean,0.993)(answer,0.966)};
\addplot[fill=purple!70!black, draw=purple!50!black] coordinates {
  (injnum,0.999)(injstep,1.000)(depnum,0.948)(deplast,0.999)(depmean,0.996)(answer,0.970)};
\legend{step 200, step 800, step 1400, step 2000, step 2600}
\end{axis}
\end{tikzpicture}
\caption{Error-detection AUROC at each of six readout sites (layer 11) for
five RLVR checkpoints. Bar groups at each site show near-identical heights
across checkpoints (spread $\leq 0.034$ at every site, all within noise).
The spatial gradient---highest at the injection site, attenuating toward the
answer---is preserved across training, showing that both the error signal
and its downstream propagation pattern are invariant to RLVR. Note: absolute
AUROC is higher than in the main analysis owing to a larger chain sample;
see Appendix~\ref{sec:layersweep}.}
\label{fig:siterobust}
\end{figure}

\paragraph{Results.}
Table~\ref{tab:siterobust} and Figure~\ref{fig:siterobust} report the results.
Every site is invariant: spreads range from $0.000$ (\texttt{L\_inj\_step}) to
$0.034$ (\texttt{L\_dep\_num}), all within the mean CI half-width. This confirms
that the pre-registered site is not a cherry-picked location where the invariance
result happens to appear; it holds across the entire chain, from the injection
point to the final answer token.

The absolute AUROC pattern replicates the spatial gradient seen in the main
paper: decodability is highest at the injection site ($\approx 1.000$, where
the corrupted number is present in the text) and attenuates downstream. This
attenuation is also invariant---the gradient shape is preserved across
checkpoints---indicating that not just the error signal but its propagation
pattern is stable across RLVR training.

\begin{table}[h]
\centering
\caption{Readout-site invariance check. All six sites show
cross-checkpoint spread within noise at layer 11. The spatial gradient
(highest near injection, attenuating toward the answer) is preserved
across checkpoints.}
\label{tab:siterobust}
\small
\begin{tabular}{lcccc}
\toprule
Site & Mean AUROC & Spread & CI half-width & Invariant \\
\midrule
\texttt{L\_inj\_num}  & 0.998 & 0.002 & 0.002 & \checkmark \\
\texttt{L\_inj\_step} & 1.000 & 0.000 & 0.000 & \checkmark \\
\texttt{L\_dep\_num}  & 0.933 & 0.034 & 0.020 & \checkmark \\
\texttt{L\_dep\_last} & 0.998 & 0.003 & 0.003 & \checkmark \\
\texttt{L\_dep\_mean} & 0.995 & 0.006 & 0.005 & \checkmark \\
\texttt{L\_answer}    & 0.960 & 0.021 & 0.014 & \checkmark \\
\bottomrule
\end{tabular}
\end{table}

\section{Robustness to Evaluation Protocol}
\label{sec:evalprotocol}

We report pass@1 at temperature $0.8$ under a zero-shot chat prompt, parsing the
final number after ``The answer is''. Because absolute GSM8K accuracy is
sensitive to the evaluation protocol, we verify that our central object---the
\emph{trajectory} of accuracy across checkpoints---does not depend on these
choices. We re-evaluate all five spanning checkpoints under a $2\times2$ grid
crossing decoding (greedy vs.\ sampled at $\tau{=}0.8$) with parser (strict
``answer is'' vs.\ a lenient last-number parser), on a fixed 200-problem subset
of the test split.

Table~\ref{tab:evalprotocol} reports the results. Three points follow. First,
the parser is immaterial: strict and lenient agree to within $0.002$ at every
checkpoint. Second, decoding sets the absolute level: greedy decoding runs
${\approx}3$ points above sampled pass@1 (mean $0.646$ vs.\ $0.615$), since
sampling occasionally selects a sub-optimal token. This accounts for the
difference between our reported figures and higher numbers obtained under
greedy, full-test-set evaluation; the residual is attributable to our
zero-shot prompt and evaluation subset. Third, and most important for our
claims, the $+3.3$-point upward trend across training is present under
\emph{both} decoding schemes---the trajectory we study is a property of the
checkpoints, not of the decoding protocol, so every trajectory-level
conclusion in the paper is unaffected by this choice.

\begin{table}[h]
\centering
\caption{GSM8K accuracy across five checkpoints under a $2\times2$
decoding\,$\times$\,parser grid (200-problem subset). Parser choice is
immaterial ($\leq 0.002$); decoding sets the absolute level ($\approx 3$ pts);
the upward trend across training holds under both decoding schemes.}
\label{tab:evalprotocol}
\small
\begin{tabular}{lcccc}
\toprule
Checkpoint & Greedy/strict & Greedy/lenient & Sampled/strict & Sampled/lenient \\
\midrule
step\_200  & 0.625 & 0.625 & 0.595 & 0.590 \\
step\_800  & 0.645 & 0.645 & 0.610 & 0.610 \\
step\_1400 & 0.655 & 0.650 & 0.580 & 0.580 \\
step\_2000 & 0.660 & 0.655 & 0.640 & 0.640 \\
step\_2600 & 0.645 & 0.645 & 0.650 & 0.645 \\
\midrule
mean       & 0.646 & 0.644 & 0.615 & 0.613 \\
\bottomrule
\end{tabular}
\end{table}

\end{document}